\documentclass{svproc}
\usepackage{url}

\usepackage{amsmath}
\usepackage{algorithmic}
\usepackage{pgf}
\usepackage{tikz}
\usetikzlibrary{arrows,shapes,automata,backgrounds,positioning}
\usepackage[caption=false]{subfig}
\usepackage{multirow}
\usepackage{url}
\usepackage{xcolor}
\usepackage{multirow}
\usepackage{amsfonts}

\usepackage{mathtools}
\usepackage{bm}
\usepackage{cite}

\usepackage{amsmath}
\usepackage{algorithmic}

\newcommand{\humanloss}[0]{0.022}
\newcommand{\bestloss}[0]{0.01701}
\newcommand{\sngnnloss}[0]{0.03173}
\newcommand{\subjects}[0]{4}
\newcommand{\bestFrameworkArchitecture}[0]{DGL/GAT}
\newcommand{\trainingSessions}[0]{5000}

\begin{document}
\mainmatter              
\title{Graph Neural Networks for Human-aware Social Navigation}
\titlerunning{Graph Neural Networks for Human-aware Social Navigation}  
%
\author{
Luis J. Manso\inst{1} \and Ronit~R.~Jorvekar\inst{2} \and Diego R. Faria\inst{1}\and \\ Pablo~Bustos\inst{3}\and Pilar~Bachiller\inst{3}
}
\authorrunning{
Manso et al.
} 

\institute{
Dept. of Computer Science, College of Engineering and Physical Sciences,\\Aston University, United Kingdom \\
\and
Dept. of Computer Engineering, Pune Institute of Computer Technology, India\\
\and
Robotics and Artificial Vision Laboratory, University of Extremadura, Spain\\
\email{l.manso@aston.ac.uk},\\
}

\maketitle              

\begin{abstract}
Autonomous navigation is a key skill for assistive and service robots.
To be successful, robots have to comply with social rules, such as avoiding the personal spaces of the people surrounding them, or not getting in the way of human-to-human and human-to-object interactions.
This paper suggests using Graph Neural Networks to model how inconvenient the presence of a robot would be in a particular scenario according to learned human conventions so that it can be used by path planning algorithms.
To do so, we propose two automated scenario-to-graph transformations and benchmark them with different Graph Neural Networks using the SocNav1 dataset~\cite{manso2020socnav}.
We achieve close-to-human performance in the dataset and argue that, in addition to its promising results, the main advantage of the approach is its scalability in terms of the number of social factors that can be considered and easily embedded in code in comparison with model-based approaches.
The code used to train and test the resulting graph neural network is available in a public repository.
\end{abstract}

\section{Introduction}\label{intro}
Human-aware robot navigation deals with the challenge of endowing mobile social robots with the capability of considering the emotions and safety of people nearby while moving around their surroundings.
There is a wide range of works studying human-aware navigation from a considerably diverse set of perspectives.
Pioneering works such as~\cite{Pacchierotti2005} started taking into account the personal spaces of the people surrounding the robots, often referred to as proxemics.
In addition to proxemics, human motion patterns were analysed in~\cite{Hansen2009} to estimate whether humans are willing to interact.
Semantic properties were also considered in~\cite{Cosley2009}.
Although not directly applied to navigation, the relationships between humans and objects were used in the context of ambient intelligence in~\cite{Bhatt2010}.
Proxemics and object affordances were jointly considered in~\cite{Vega2019} for navigation purposes.
Two extensive surveys on human-aware navigation can be found in~\cite{Rios-Martinez2015} and~\cite{Charalampous2017}.
\par

Despite the previously mentioned approaches being built on well-studied psychological models, they have limitations.
Considering new factors programmatically (\textit{i.e.}, writing additional code) involves a potentially high number of coding hours, makes systems more complex, and increases the chances of including bugs.
Additionally, with every new aspect to be considered for navigation, the decisions made become less \textit{explainable}, which is precisely one of the main advantages of model-based approaches over data-driven ones.
Besides the mentioned model scalability and explainability issues, model-based approaches have the intrinsic and rather obvious limitation that they only account for what the model explicitly considers.
Given that these models are manually written by humans, they cannot account for aspects that the designers are not aware of.
\par

Approaches leveraging machine learning have also been published.
The parameters of a social force model (see~\cite{Helbing1995}) are learned in~\cite{Ferrer2013} and~\cite{Patompak2019} to navigate in human-populated environments.
Inverse reinforcement learning is used in~\cite{Ramon-Vigo2014} and~\cite{Vasquez2014} to plan navigation routes based on a list of humans in a radius.
Social norms are implemented using deep reinforcement learning in~\cite{Chen2017}, again, considering a set of humans.
An approach modelling crowd-robot interaction and navigation control is presented in~\cite{Chen2019}.
It features a two-module architecture where single interactions are modelled and then aggregated.
Although its authors reported good qualitative results, the approach does not contemplate integrating additional information (\textit{e.g.}, relations between humans and objects, structure and size of the room).
The work in~\cite{martins2019clusternav} tackles the same problem using Gaussian Mixture Models.
It has the advantage of requiring less training data, but the approach is also limited in terms of the input information used.
\par

All the previous works and many others not mentioned have achieved outstanding results.
Some model-based approaches such as~\cite{Cosley2009} or~\cite{Vega2019} can leverage structured information to take into account space affordances.
Still, the data considered to make such decisions are often handcrafted features based on an arbitrary subset of the data that a robot would be able to work with.
There are many reasons motivating to seek a learning-based approach not requiring feature handcrafting or manual selection.
Their design is time-consuming and often requires a deep understanding of the particular domain (see discussion in~\cite{Lecun2015}).
Additionally, there is generally no guarantee that a particular hand-engineered set of features is close to being the best possible one.
On the other hand, most end-to-end deep learning approaches have important limitations too.
They require a big amount of data and computational resources that are often scarce and expensive, and they are hard to explain and manually fine-tune.
Somewhere in the middle of the spectrum, we have proposals advocating not to choose between hand-engineered features or end-to-end learning.
In particular, \cite{Battaglia2018} proposes Graph Neural Networks (GNNs) as a means to perform learning that allows combining raw data with hand-engineered features, and most importantly, learn from structured information.
The relational inductive bias of GNNs is specially well-suited to learn about structured data and the relations between different types of entities, often requiring less training data than other approaches.
In this line, we argue that using GNNs for human-aware navigation makes possible integrating new social cues in a straightforward fashion, by including more data in the graphs that they are fed.
\par

In this paper, we use different GNN models to estimate social navigation compliance, \textit{i.e.}, given a robot pose and a scenario where humans and objects can be interacting, estimating to what extent a robot would be disturbing the humans if it was located in such a pose.
GNNs are proposed because the information that social robots can work with is not just a map and a list of people, but a more sophisticated data structure where the entities represented have different relations among them.
For example, social robots can have information about who a human is talking to, where people are looking at, who is friends with who, or who is the owner of an object in the scenario.
Regardless of how this information is acquired, it can be naturally represented using a graph, and GNNs are a particularly well-suited and scalable machine learning approach to work with these graphs.
\section{Graph neural networks}
\begin{figure*}[ht]
 \centering
 \includegraphics[width=0.55\textwidth,keepaspectratio=true]{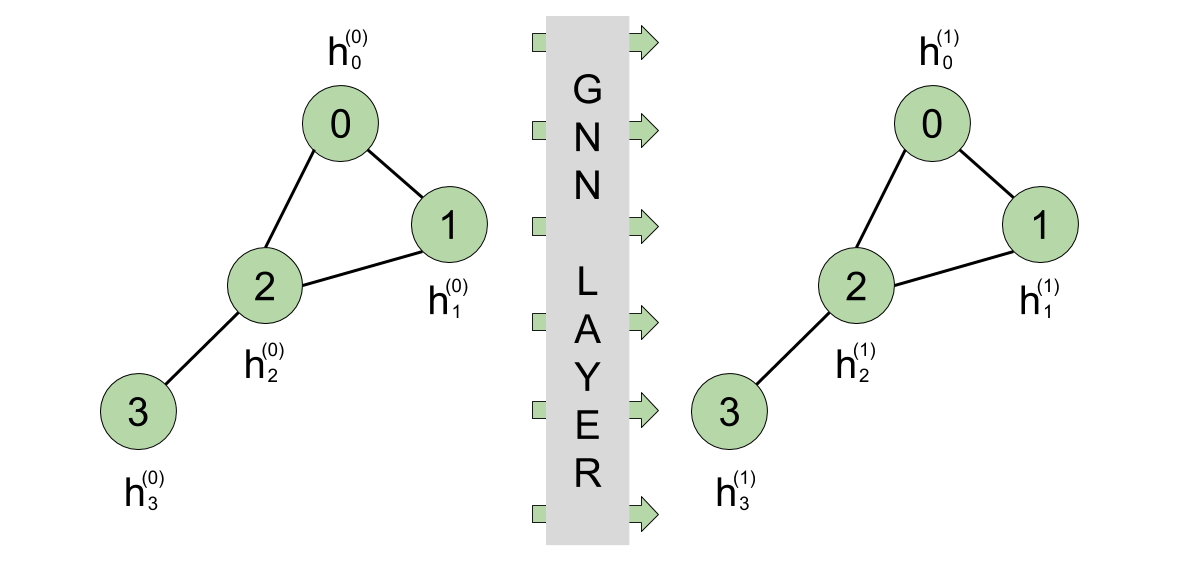}
 \caption{A basic GNN block/layer. GNN layers output updated versions of the input graph. These updated graphs have the same nodes and links, but the feature vectors of the nodes will generally differ in size and content depending on the feature vectors of their neighbours and their own vectors in the input graph. A GNN is usually composed of several stacked GNN layers. Higher level features are learnt in the deeper layers, to that the output of any of the nodes in the last layer can used for classification or regression purposes.}
 \label{fig:gnnlayer}
\end{figure*}

Graph Neural Networks (GNNs) are a family of machine learning approaches based on neural networks that take graph-structured data as input.
They allow classifying and making regressions on graphs, nodes, edges, as well as predicting link existence when working with partially observable phenomena.
Except for few exceptions (\textit{e.g.},~\cite{ying2018hierarchical}) GNNs are composed by similar stacked blocks/layers operating on a graph whose structure remains static but the features associated to its nodes are updated in every layer of the network (see Fig.~\ref{fig:gnnlayer}).
\par
As a consequence, the features associated to the nodes of the graph in each layer become more abstract and are influenced by a wider context as layers go deeper.
The features in the nodes of the last layer are frequently used to perform the final classification or regression.
\par

The first published efforts on applying neural networks to graphs date back to~\cite{Sperduti1998}.
GNNs were further studied and formalised in~\cite{Gori2005} and~\cite{F.2009}.
However, it was with the appearance of Gated Graph Neural Networks (GG-NNs, \cite{Li2015}) and especially Graph Convolutional Networks (GCNs, \cite{Kipf2016a}) that GNNs gained traction.
The work presented in~\cite{Battaglia2018} reviewed and unified the notation used in the GNNs existing to the date.
\par

Graph Convolutional Networks (GCN)~\cite{Kipf2016a} are one of the most common GNN blocks.
Because of its simplicity, we build on the GCN block to provide the reader with an intuition of how GNNs work in general.
Following the notation proposed in~\cite{Battaglia2018}, GCN blocks operate over a graph $G=(V,E)$, where $V=\{v_i\}_{i=1:N^v}$ is a set of nodes, being $v_i$ the feature vector of node $i$ and $N^v$ the number of vertices in the graph.
$E$ is a set of edges $E=\{(s_k,r_k)\}_{k=1:N^e}$, where $s_k$ and $r_k$ are the source and destination indices of edge $k$ and $N^e$ is the number of edges in the graph.
Each GCN layer generates an updated representation $v_i'$ for each node $v_i$ using two functions:
$$\overline{e}_i = \displaystyle\rho^{e \rightarrow v}(E) = \sum_{\{k:r_k=i\}}e_k,$$
$$v_i' = \phi^{v}(\overline{e}_i, v_i)=NN_v([\overline{e}_i,v_i]).$$
For every node $v_i$, the first function ($\displaystyle\rho^{e \rightarrow v}(E)$) aggregates the feature vectors of other nodes with an edge towards it and generates a temporary aggregated feature $\overline{e}_i$ which is used by the second function.
In a second pass, the function $\phi^{v}(\overline{e}_i, v_i)$ is used to generate updated $v_i'$ feature vectors from the aggregated feature vectors using a neural network (usually a multi-layer perceptron, but the framework does not make any assumption on this).
Such a learnable function is the same for all the nodes. 
By stacking several blocks where features are aggregated and updated, the feature vectors can carry information from nodes far away in the graph and convey higher level features that can be finally used for classification or regressions.
\par

Several means of improving GCNs have been proposed.
Relational Graph Convolutional Networks (RGCNs~\cite{Schlichtkrull2018}) extends GCNs by considering different types of edges separately and applies the resulting model to vertex classification and link prediction. 
Graph Attention Networks (GATs~\cite{Velickovic2018}) extend GCNs by adding self-attention mechanisms (see~\cite{vaswani2017attention}) and applies the resulting model to vertex classification.
For a more detailed review of GNNs and the generalised framework, please refer to~\cite{Battaglia2018}.
\par

\section{Formalisation of the problem} \label{problem}
The aim of this work is to analyse the scope of GNNs in the field of human-aware social navigation.
Our study has been set up using the \textit{SocNav1} dataset (see~\cite{manso2020socnav}) which provides social compliance labels for specific scenarios.
It contains scenarios with a robot in a room, a number of objects and a number of people that can potentially be interacting with other objects or people.
Each sample is labelled with a value from $0$ to $100$ depending on to what extent the subjects that labelled the scenarios considered that the robot is disturbing the people in the scenario.
The dataset provides 16336 labelled samples to be used for training purposes, 556 scenarios as the development dataset and additional 556 for final testing purposes.

As previously noted, GNNs are a flexible framework that allows working somewhere in the middle of end-to-end and feature engineered learning.
Developers can use as many data features as desired and are free to structure the input graph data as they please.
The only limitations are those of the particular GNN layer blocks used.
In particular, while GCN and GAT do not support labelled edges, RGCN and GG-NN do.
To account for this limitation, two automated scenario-to-graph transformations were used in the experiments, depending on the GNN block to be tested: one without edge labels and one with them.

The first version of the scenario-to-graph transformation used to represent the scenarios does not use labelled edges.
It uses 6 node types (the features associated to each of the types are detailed later in this section):
\begin{itemize}
 \item \textbf{robot:} The dataset only includes one robot in each scenario, so there is just one robot symbol in each of the graphs. However, GNNs do not have such restriction.
 \item \textbf{wall:} A node for each of the segments defining the room. They are connected to the room node.
 \item \textbf{room:} Used to represent the room where the robot is located. It is connected to the robot.
 \item \textbf{object:} A node for each object in the scenario.
 \item \textbf{human:} A node for each human. Humans might be interacting with objects or other humans.
 \item \textbf{interaction:} An interaction node is created for every human-to-human or human-to-object interaction.
\end{itemize}
\par

\begin{figure*}[!ht] 
 \centering
\subfloat[{An area of a scenario where two humans interacting in a room are depicted.}]{\includegraphics[width=0.42\textwidth,keepaspectratio=true]{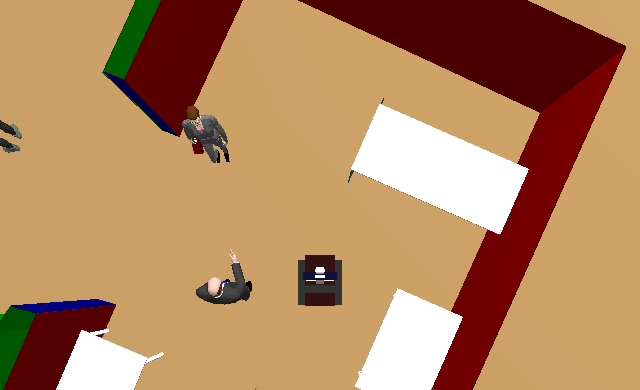} \label{fig:scenario1}}
 \hspace{0.4cm}
\subfloat[Heat map of the social compliance estimation for the area shown in Fig.~\ref{fig:scenario1} in the different positions in the environment.]{ \includegraphics[width=0.42\textwidth,keepaspectratio=true]{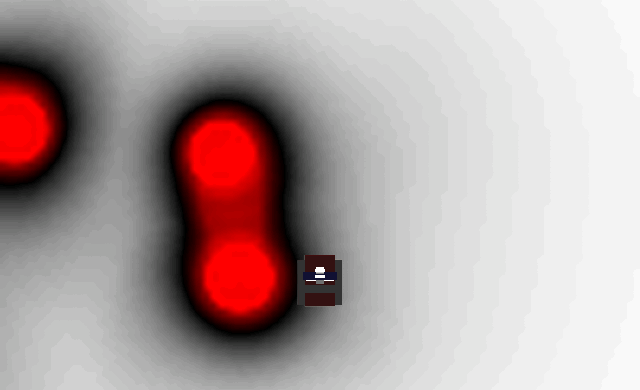}\label{fig:scenario1_hm}} \\
\subfloat[An area of the same scenario with two humans. The human on the left is not engaged in any interaction. The human on the right is interacting with the object in front of her.]{\includegraphics[width=0.42\textwidth,keepaspectratio=true]{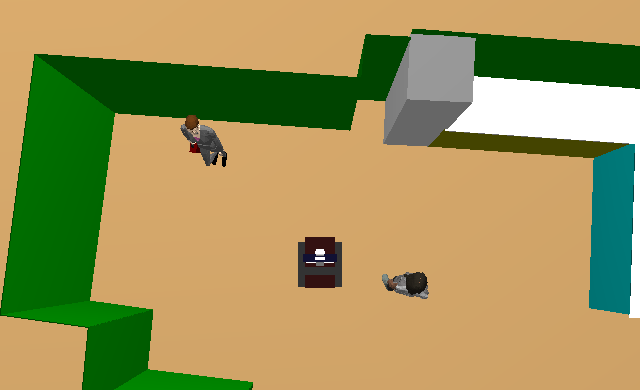} \label{fig:scenario2}}
 \hspace{0.4cm}
\subfloat[Heat map of the social compliance estimation for the area shown in Fig.~\ref{fig:scenario2} in the different positions in the environment.]{ \includegraphics[width=0.42\textwidth,keepaspectratio=true]{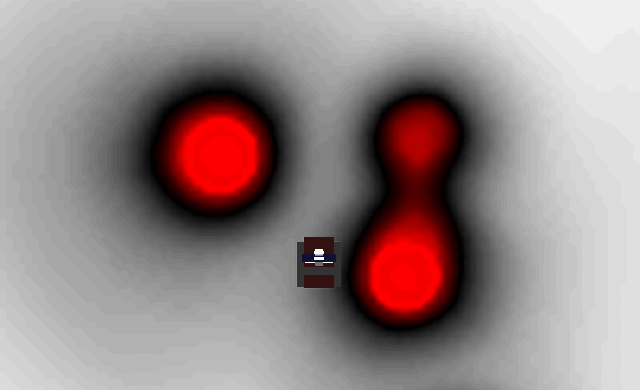}\label{fig:scenario2_hm}}

\caption{Different areas of a scenario where social interactions are being held and their corresponding estimated heat map of \textit{``social inconvenience''}.}
\label{fig:scenarios}
\end{figure*}

\begin{figure*}[!t]
\centering
\hspace{1cm}
\subfloat[{Graph without labelled edges.}]{\begin{tikzpicture}[font=\small,->,>=stealth',shorten >=1pt,auto,semithick]
  \tikzstyle{every state}=[fill=lightgray,draw=black,text=black]
  \node[state](robot)               {$robot$};
  \node[state](room) [above left=15mm and 4mm of robot] {$room$};  
  \node[state](w1)   [above right=2mm and 17mm of room] {$w_1$};
  \node[state](w2)   [below right=2mm and 17mm of room] {$w_2$};
  \node[state](w3)   [below left=2mm and 13mm of room] {$w_3$};
  \node[state](w4)   [above left=2mm and 13mm of room]  {$w_4$};  \node[state](h1)   [below left=-5mm and 25mm of robot]    {$h_1$};
  \node[state](h2)   [below left=18mm and 12mm of robot]    {$h_2$};
  \node[state](i1)   [below left=8mm and 21mm of robot]    {$i_1$};
  \node[state](h3)   [below right=18mm and 12mm of robot]    {$h_3$};
  \node[state](i2)   [below right=8mm and 21mm of robot]    {$i_2$};
  \node[state](o1)   [below right=-5mm and 25mm of robot]    {$o_1$};
  \node[state](h4)   [below=16mm of robot]    {$h_4$};
  \path (w1) edge              node {} (room)
        (w2) edge              node {} (room)
        (w3) edge              node {} (room)
        (w4) edge              node {} (room)
        (room) edge            node {} (robot)
        (h1) edge              node {} (robot)
        (h2) edge              node {} (robot)
        (h3) edge              node {} (robot)
        (h4) edge              node {} (robot)
        (o1) edge              node {} (robot)
        (h1) edge              node {} (i1)
        (h2) edge              node {} (i1)
        (i1) edge              node {} (h1)
        (i1) edge              node {} (h2)
        (h3) edge              node {} (i2)
        (o1) edge              node {} (i2)
        (i2) edge              node {} (o1)
        (i2) edge              node {} (h3);
\end{tikzpicture}
\label{fig:graph4}}
\hfill
\subfloat[{Graph with labelled edges.}]{\begin{tikzpicture}[font=\small,->,>=stealth',shorten >=1pt,auto,semithick]
  \tikzstyle{every state}=[fill=lightgray,draw=black,text=black]
  \node[state](robot)               {$robot$};
  \node[state](room) [above left=15mm and 4mm of robot] {$room$};  
  \node[state](w1)   [above right=2mm and 17mm of room] {$w_1$};
  \node[state](w2)   [below right=2mm and 17mm of room] {$w_2$};
  \node[state](w3)   [below left=2mm and 13mm of room] {$w_3$};
  \node[state](w4)   [above left=2mm and 13mm of room]  {$w_4$};
  \node[state](h1)   [below left=-5mm and 25mm of robot]    {$h_1$};
  \node[state](h2)   [below left=18mm and 12mm of robot]    {$h_2$};
  \node[state](h3)   [below right=18mm and 12mm of robot]    {$h_3$};
  \node[state](o1)   [below right=-5mm and 25mm of robot]    {$o_1$};
  \node[state](h4)   [below=16mm of robot]    {$h_4$};
  \path (w1) edge              node {w-r} (room)
        (w2) edge              node {w-r} (room)
        (w3) edge              node {w-r} (room)
        (w4) edge              node {w-r} (room)
        (room) edge            node {r-R} (robot)
        (h1) edge              node {h-R} (robot)
        (h2) edge              node {h-R} (robot)
        (h3) edge              node {h-R} (robot)
        (h4) edge              node {h-R} (robot)
        (o1) edge              node {o-R} (robot)
        (h1) edge              node {h-h} (h2)
        (h2) edge              node {h-h} (h1)
        (h3) edge              node {h-o} (o1)
        (o1) edge              node {o-h} (h3);
\end{tikzpicture}
\hspace{1cm}
\label{fig:graphR}}
\caption{Examples of how the scenario-to-graph transformation work based on the scenario depicted in Fig.\ref{fig:scenarios}.}
\label{fig:graphs}
\end{figure*}
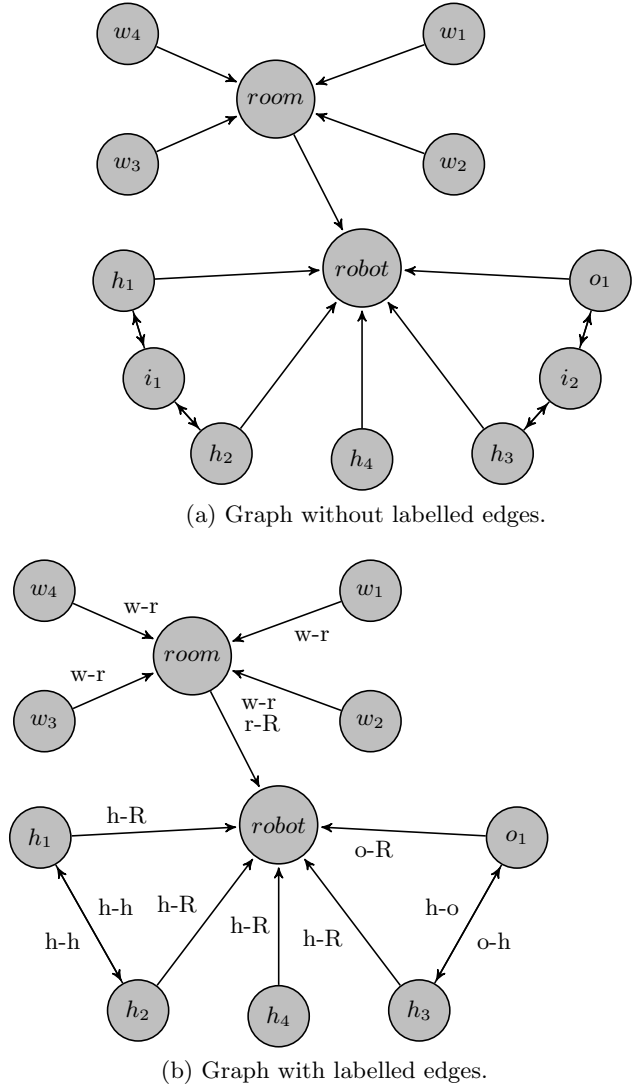

Figure~\ref{fig:scenarios} depicts two areas of a scenario where four humans are shown in a room with several objects.
Two of the humans are interacting with each other, another human is interacting with an object, and the remaining human is not engaging in interaction with any human or object.
The structure of the resulting non-labelled graph is shown in Fig.\ref{fig:graph4}.
\par

The features used for \textbf{human} and \textbf{object} nodes are: distance, the relative angle from the robot's point of view, and its orientation, from the robot's point of view too.
For \textbf{room} symbols the features are: the distance to the closest human and the number of humans.
For the \textbf{wall} segments and the \textbf{interaction} symbols, the features are the distance and orientation from the robot's frame of reference.
For wall segments, the position is the centre of the segment and the orientation is the tangent.
These features geometrically define the room, which is relevant to characterise the density of people in the room and the distance from the robot to each of the walls.
For interactions, the position is the midpoint between the interacting symbols, and the orientation is the tangent of the line connecting the endpoints.
Features related to distances are expressed in meters, whereas those related to angles are actually expressed as two different numbers, $sin(\alpha)$ and $cos(\alpha)$.
The final features of the nodes are built by concatenating the one-hot encoding that determines the type of the symbol and the features for the different node types.
It is worth noting that by building feature vectors this way, their size increases with every new type.
This limitation is currently being studied by the GNN scientific community.
\par

For the GNN blocks that can work with labelled edges, a slightly different version of the scenario-to-graph transformation is used.
The first difference is that in this version of the scenario-to-graph model there are no interaction nodes.
The elements interacting are linked to each other directly.
Robot, room, wall, human and object nodes are attributed with the same features as in the previous model.
The second difference is related to the labelling of the edges.
In this domain, the semantics of the edges can be inferred from the types of the nodes being connected.
For example, \textit{wall} and \textit{room} nodes are always connected by the same kind of relation ``composes''.
Similarly, \textit{humans} and \textit{object} nodes are always connected by the relation ``interacts\_with\_human''.
The same holds the other way around: ``composes'' relations only occur between \textit{wall} and \textit{room} nodes, and ``interacts\_with\_human'' relations only occur with \textit{humans} and \textit{object} nodes.
Therefore, for simplicity, the label used for the edges is the concatenation of the types involved.
The structure of the resulting labelled graph for the scenario depicted in Fig.\ref{fig:scenarios} is shown in Fig.\ref{fig:graphR}.

\par

Because all nodes are connected to the robot, the GNNs were trained to perform the regression on the feature vector of the \textit{robot} node in the last layer.
\par

Using the previously mentioned dataset and the two proposed scenario-to-graph transformations, different architectures using different GNN blocks were compared.

\section{Experimental results}
The experimental results are aligned with the contributions of the paper, which deal with modelling social conventions. Therefore, we assume that we can build on top of a third party body tracker and a path planning system and proceed with the evaluation of the algorithm against the dataset.
\par

The four GNN blocks used in the experiments were:
\begin{itemize}
 \item Graph Convolutional Networks (GCN)~\cite{Kipf2016a}.
 \item Gated Graph Neural Networks (GG-NN)~\cite{Li2015}.
 \item Relational Graph Convolutional Networks (RGCNs)~\cite{Schlichtkrull2018}.
 \item Graph Attention Networks (GAT)~\cite{Velickovic2018}.
\end{itemize}
\par

If available, each of these GNN blocks was benchmarked using Deep Graph Library (DGL)~\cite{wang2019dgl} and PyTorch-Geometric (PyG)~\cite{fey2019pyg}.
In addition to using several layers of the same GNN building blocks, alternative architectures combining RGCN and GAT layers were also tested:
\begin{enumerate}
 \item A sequential combination of $n$ RGCN layers followed by $n$ GAT layers with the same number of hidden units (alternative 8 in table~\ref{tbl:experiment_results}).
 \item An interleaved combination of $n$ RGCN and $n$ GAT layers with the same number of hidden units (alternative 9 in table~\ref{tbl:experiment_results}).
 \item A sequential combination of $n$ RGCN layers followed by $n$ GAT layers with a linearly decreasing number of hidden units (alternative 10 in table~\ref{tbl:experiment_results}).
\end{enumerate}
\par

As a result, 10 framework-architecture combinations were benchmarked.
Table~\ref{tbl:experiment_results} describes them and provides their corresponding performance on the development dataset.
To benchmark the different architectures, \trainingSessions{}  training sessions were launched using the SocNav1 training dataset and evaluated using the SocNav1 development dataset.
The hyperparameters were randomly sampled from the range values shown in Table~\ref{tbl:experiment_hp}.

\begin{table}[h]
\centering
\begin{tabular}{|l|l|c|c|}
\hline
\multirow{2}{*}{Alt. \#}
  & \multirow{2}{*}{Framework}
                         & Network       & Training Loss              \\ 
  &                      & architecture  & (MSE)             \\ \hline
 1& DGL                  & GCN           & 0.02283           \\ \hline
 2& \textbf{DGL}         & \textbf{GAT}  & \textbf{0.01701}  \\ \hline
 3& DGL                  & GAT2          & 0.01740                  \\ \hline
 4& PyG                  & GCN           & 0.29778           \\ \hline
 5& PyG                  & GAT           & 0.01804           \\ \hline
 6& PyG                  & RGCN          & 0.02827           \\ \hline
 7& PyG                  & GG-NN         & 0.02718           \\ \hline
 8& PyG                  & RGCN$||$GAT 1 & 0.02238           \\ \hline
 9& PyG                  & RGCN$||$GAT 2 & 0.02147           \\ \hline
10& PyG                  & RGCN$||$GAT 3 & 0.0182           \\ \hline
\end{tabular}
\vspace{1mm}
\caption{A description of the different framework/architecture combinations and the experimental results obtained from their benchmark for the SocNav1 dataset.}
\label{tbl:experiment_results}
\end{table}

\begin{table}[h]
\centering
\begin{tabular}{|l|c|c|}
\hline
Hyperparameter & Min  & Max   \\ \hline
epochs         & \multicolumn{2}{c|}{1000}  \\ \hline
patience       & \multicolumn{2}{c|}{5}     \\ \hline
batch size     & 100  & 1500  \\ \hline
hidden units   & 50   & 320   \\ \hline
attention heads& 2    & 9     \\ \hline
learning rate  & 1e-6 & 1e-4  \\ \hline
weight decay   & 0.0  & 1e-6  \\ \hline
layers         & 2    & 8      \\ \hline
dropout        & 0.0  & 1e-6  \\ \hline
alpha          & 0.1  & 0.3   \\ \hline
\end{tabular}
\vspace{1mm}
\caption{Hyperparameter values. Only applicable to Graph Attention Network blocks.}
\label{tbl:experiment_hp}
\end{table}

The results obtained (see table~\ref{tbl:experiment_results})
show that, for the dataset and the framework/architecture combinations benchmarked, \bestFrameworkArchitecture{} delivered the best results, with a training loss of \bestloss{} for the development dataset.
The parameters used by the \bestFrameworkArchitecture{} combination were: batch size: 273, number of hidden units: 129, number of attention heads: 2, number of attention heads in the last layer: 3, learning rate: 5e-05, weight decay regularisation: 1e-05, number of layers: 4, no dropout, alpha parameter of the ReLU non-linearity 0.2114.
After selecting the best set of hyperparameters, the network was compared with a third \textit{test} dataset, \textbf{obtaining an MSE of \sngnnloss{}}.
Figures~\ref{fig:scenario1_hm} and~\ref{fig:scenario2_hm} provide an intuition of the output of the network for the scenarios depicted in figures~\ref{fig:scenario1} and~\ref{fig:scenario2} considering all the different positions of the robot in the environment when looking along the $y$ axis.
\par

\begin{figure}[!t]
\centering
\includegraphics[width=0.6\columnwidth]{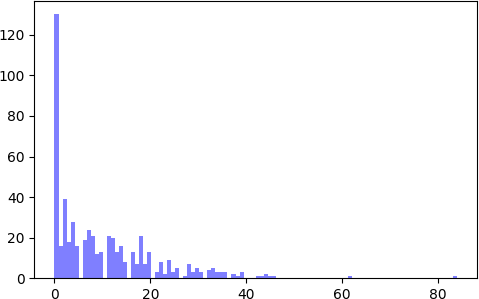}
\caption{Histogram of the absolute error in the test dataset for the network performing best in the development dataset.}
\label{fig:histogram}
\end{figure}

It is worth noting that, due to the subjective nature of the labels in the dataset (human feelings are utterly subjective), there is some level of disagreement even among humans.
To compare the performance of the network with human performance, we asked \subjects{} subjects to label all the scenarios of the development dataset.
The mean MSE obtained for the different \textbf{subjects was \humanloss{}}.
This means that the network performs close to human accuracy.
Figure~\ref{fig:histogram} shows an histogram comparing the error made by the GNN-based regression in comparison to humans.
\par

Most algorithms presented in section~\ref{intro} deal with modelling human intimate, personal, social and interaction spaces instead of social inconvenience, which seems to be a more general term.
Keeping that in mind, the algorithm proposed in~\cite{Vega2019} was tested against the test dataset and got a MSE of 0.12965.
The relative bad performance can be explained by the fact that other algorithms do not take into account walls and that their actual goal is to model personal spaces instead of feelings in general.
\par

Regarding the effect of the presence of humans, we can see from Fig.~\ref{fig:scenario2_hm} that the learnt function is slightly skewed to the front of the humans, but not as much as modelled in other works such as~\cite{kirby2009companion} or~\cite{Vega2019}.
One of the possible reasons why the ``personal space'' is close to being circular is the fact that, in the dataset, humans appear to be standing still.
It is still yet to be studied, probably using more detailed and realistic datasets, how would the personal space look like if humans were moving.
\par

The results obtained using GNN blocks supporting edge labels were inferior to those obtained using GAT, which does not support edge labels.
Two reasons might be the cause of this phenomena: \textbf{a)}~as mentioned in section~\ref{problem} the labels edges can be inferred from the types of the nodes, so that information is to some extent redundant; \textbf{b)}~the inductive bias of GATs is strong and appropriate for the problem at hand.
This does not mean that the same results would be obtained in other problems where the label of the edges cannot be inferred.
\par

\section{Conclusions}
Although there have been attempts to predict people's trajectories using graph neural networks, to our knowledge, this paper presented the first graph neural network modelling human-aware navigation conventions.
The scenario-to-graph transformation model and the graph neural network developed as a result of the work presented in this paper achieved a performance comparable to that of humans.
The final model can be used as part of a social navigation system to predict the degree of acceptance of the robot position according to social conventions.
Thus, given the graph representation of a scenario, the GNN provides a quantification of how good a specific robot location is for humans.
Even though the results achieved are remarkable, the key fact is that this approach allows to include more relational information.
This will allow to include more sources of information in our decisions without a big impact in the development.
\par

There is room for improvement, particularly related to: \textbf{a)}~personalisation (different people generally feel different about robots), and \textbf{b)}~movement (the inconvenience of the presence of a robot is probably influenced by the movement of the people and the robot).
Still, we include interactions and walls, features which are seldom considered in other works.
As far as we know, interactions have only been considered in~\cite{Vega2019} and \cite{Cruz-maya2019}.
\par

The code to test the resulting GNN has been published in a public repository as open-source software
\footnote{https://github.com/robocomp/sngnn}, as well as the code implementing the scenario-to-graph transformation and the code to train the models suggested.
\par

%
%
\bibliographystyle{spmpsci_unsrt} 
\bibliography{bib}

\end{document}